\documentclass[lettersize,journal]{IEEEtran}
\usepackage{amsmath,amsfonts}
\usepackage{algorithmic}
\usepackage{algorithm}
\usepackage{array}
\usepackage[caption=false,font=normalsize,labelfont=sf,textfont=sf]{subfig}
\usepackage{textcomp}
\usepackage{stfloats}
\usepackage{url}
\usepackage{verbatim}
\usepackage{graphicx}
\usepackage{cite}

\usepackage{xspace}
\usepackage{xcolor}
\usepackage{mwe}
\usepackage{booktabs}
\usepackage{multirow}
\usepackage{makecell} 
\usepackage{array}    
\usepackage{diagbox}
\usepackage{enumerate}
\usepackage{microtype}

\usepackage[textsize=tiny]{todonotes}
\begin{document}

\title{BOSS: Benchmark for Observation Space \\ Shift in Long-Horizon Task}

\author{
    Yue Yang\textsuperscript{1}, 
    Linfeng Zhao\textsuperscript{2}, 
    Mingyu Ding\textsuperscript{1}, 
    Gedas Bertasius\textsuperscript{1}, 
    Daniel Szafir\textsuperscript{1}%
    \thanks{This work has been submitted to the IEEE for possible publication. Copyright may be transferred without notice, after which this version may no longer be accessible.}
    \thanks{\textsuperscript{1} The University of North Carolina at Chapel Hill.}%
    \thanks{\textsuperscript{2} Northeastern University.}%
    \thanks{Contact Email: \texttt{yygx@cs.unc.edu}}%
}



\definecolor{mygreen}{HTML}{00B050}
\definecolor{myred}{HTML}{FF5751}
\definecolor{myblue}{HTML}{4472c4}

\newcommand{\pb}{OSS\xspace}
\newcommand{\bm}{BOSS\xspace}
\newcommand{\bma}{BOSS-C1\xspace}
\newcommand{\bmb}{BOSS-C2\xspace}
\newcommand{\bmc}{BOSS-C3\xspace}
\newcommand{\bl}{BaselineName\xspace}
\newcommand{\taskoriginal}{\bm-44\xspace}

\newcommand{\my}[1]{\textcolor{red}{[Mingyu: #1]}}
\newcommand{\gb}[1]{{\textcolor{blue}{GB: #1}}}
\newcommand{\yy}[1]{{\textcolor{green}{(YY: #1)}}}
\newcommand{\ds}[1]{{\textcolor{yellow}{DS: #1}}}

\newcommand{\todoinline}[1]{\todo[inline, size=\small, color=orange!50]{#1}}
\newcommand{\todozlf}[2][]{\todo[color=yellow!80!black, #1]{LZ: #2}}
\newcommand{\notezlf}[1]{\todo[color=blue!50!white]{[LZ] #1}}
\newcommand{\todoilzlf}[2][]{\todo[inline, size=\small, color=yellow!80!black, #1]{LZ: #2}}
\newcommand{\todonote}[1]{\todo[color=blue!50!white]{#1}}
\newcommand{\todoilnote}[1]{\todo[inline, size=\small, color=blue!50!white]{#1}}

\maketitle



\begin{abstract}



Robotics has long sought to develop visual-servoing robots capable of completing previously unseen long-horizon tasks. Hierarchical approaches offer a pathway for achieving this goal by executing skill combinations arranged by a task planner, with each visuomotor skill pre-trained using a specific imitation learning (IL) algorithm. However, even in simple long-horizon tasks like skill chaining, hierarchical approaches often struggle due to a problem we identify as \textbf{Observation Space Shift} (\pb), where the sequential execution of preceding skills causes shifts in the observation space, disrupting the performance of subsequent individually trained skill policies. To validate \pb and evaluate its impact on long-horizon tasks, we introduce \bm (a \underline{B}enchmark for \underline{O}bservation \underline{S}pace \underline{S}hift). \bm comprises three distinct challenges: ``Single Predicate Shift'', ``Accumulated Predicate Shift'', and ``Skill Chaining'', each designed to assess a different aspect of \pb's negative effect. We evaluated several recent popular IL algorithms on \bm, including three Behavioral Cloning methods and the Visual Language Action model OpenVLA. Even on the simplest challenge, we observed average performance drops of 67\%, 35\%, 34\%, and 54\%, respectively, when comparing skill performance with and without \pb. 
Additionally, we investigate a potential solution to \pb that scales up the training data for each skill with a larger and more visually diverse set of demonstrations, with our results showing it is not sufficient to resolve \pb.
The project page is: \texttt{https://boss-benchmark.github.io/}


\end{abstract}


\vspace{-1.2em}

\section{Introduction}
\label{sec:introduction}


Recent advances in robotic learning have demonstrated its potential in diverse manipulation applications~\cite{ravichandar2020recent}, including manufacturing~\cite{liu2022robot}, sports~\cite{zaidi2023athletic, chen2021learning}, and household tasks~\cite{yang2024enhancing, yang2024arcade}. Imitation Learning (IL), which empowers end users to teach robot skills and behaviors through demonstrations, has become a prevalent approach in developing various skill controllers~\cite{chi2023diffusion, kim2024openvla}. However, IL algorithms remain limited to relatively short-horizon tasks due to covariate shift~\cite{ross2011reduction, chang2021mitigating}, where small discrepancies in action predictions accumulate over time. This challenge becomes even more pronounced in long-horizon tasks, as such errors propagate across multiple sequential steps, creating a significant barrier to achieving general-purpose robots. Task and Motion Planning (TAMP)~\cite{garrett2021integrated, guo2023recent} and recent advancements in large language model based planning~\cite{ahn2022can, singh2023progprompt, du2023video} offer a solution by integrating IL into a hierarchical framework~\cite{xu2023xskill, zhu2022bottom, shiarlis2018taco}, decomposing end-to-end long-horizon IL into high-level task planning and low-level visuomotor skill execution.


\begin{figure}[t]
    \centering
    \includegraphics[width=0.5\textwidth]{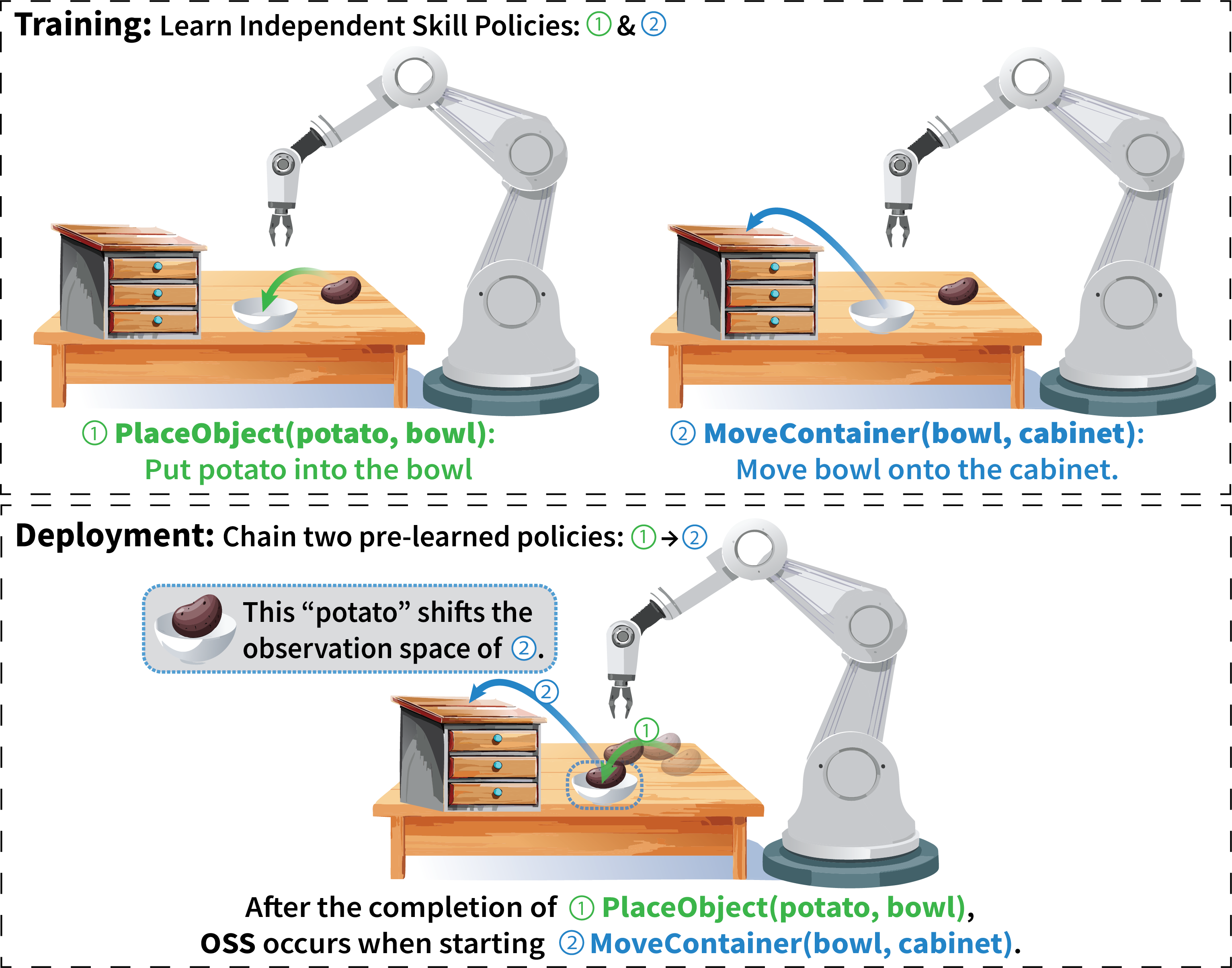} 
    \caption{The example illustrates how Observation Space Shift (\pb) occurs when chaining two pre-trained skills: \texttt{PlaceObject(potato, bowl)} and \texttt{MoveContainer(bowl, cabinet)}. During deployment, \pb arises in \texttt{MoveContainer(bowl, cabinet)} because the observation space changes, with a potato inside the bowl instead of the empty bowl scenario from \texttt{MoveContainer(bowl, cabinet)}'s pre-training. 
    }
    \vspace{-2em}
    \label{fig:osm_illustration}
\end{figure}


Skill chaining, the sequential execution of pre-learned skills, is a simple yet powerful structure for tackling complex long-horizon tasks~\cite{bagaria2019option, konidaris2009skill}.
However, failures often arise in a skill chain when a skill encounters initial states that were not seen during training~\cite{lee2019composing, clegg2018learning}. 
Specifically, the terminal state of a preceding skill may fall outside the initial state distribution the next skill policy was trained on, leading to task failure. For instance, consider a household long-horizon task illustrated in Fig.~\ref{fig:osm_illustration}, which involves two skills: \texttt{PlaceObject(potato, bowl)} and \texttt{MoveContainer(bowl, cabinet)}.
When training the visuomotor policy for \texttt{MoveContainer(bowl, cabinet)}, the initial state set excludes visual states (e.g., images) containing anything inside the bowl.
However, the terminal state of \texttt{PlaceObject(potato, bowl)} places a potato in the bowl, creating a visual mismatch that can cause the \texttt{MoveContainer(bowl, cabinet)} policy to fail. We refer to this issue as \textbf{Observation Space Shift (\pb)}, a problem that frequently arises during skill transitions in visual-input long-horizon tasks. A formal definition of \pb is provided in Section~\ref{sec:problem_definition}.

As a result, ensuring smooth skill transitions is critical in skill chaining and serves as a key factor for completing long-horizon tasks~\cite{li2024auxiliary}.
Previous researchers tackle the skill transition problem through two main strategies. The first line of offline approaches involves learning a transition policy to shift the state from the terminal state of one skill to the initial state of the next~\cite{watahiki2022one, byun2021training, lee2019composing}.
The second approach ensures that the initial state set of the next skill matches the terminal state set of the previous skill by fine-tuning the policy of the preceding skill, the next skill, or both in an online manner~\cite{li2024auxiliary, chen2023sequential, lee2021adversarial}. However, all these previous works design observation spaces that exclude visual inputs,
causing the above methods, whether using offline or online learning, to rely on assumptions that are often impractical for visuomotor IL policies. 1) Offline methods assume that the initial states for each skill are always reachable, which is often unreasonable for visual-input long-horizon tasks, as it may require undoing previously completed skills. For example, as shown in Figure~\ref{fig:osm_illustration}, offline methods would need to learn a transition policy to remove the potato from the bowl in order to enable the observation space to match the initial state of \texttt{MoveContainer(bowl, cabinet)} (i.e., an empty bowl), which is logically invalid. 2) Online approaches, on the other hand, assume that policies can be continuously fine-tuned to handle new initial states. This assumption is equally problematic for visual-input long-horizon tasks, where \pb is highly likely to happen at each skill transition. The frequent occurrence of \pb necessitates repeated fine-tuning, leading to substantial computational overhead and inefficiency.

These limitations motivate us to propose a dedicated \textbf{Benchmark for Observation Space Shift (\bm)}, to demonstrate the significant negative impact \pb has on long-horizon tasks, facilitate a deeper understanding of the problem, and inspire potential solutions (Section~\ref{sec:bm_challenge}). Built on the LIBERO simulator~\cite{liu2024libero}, the benchmark consists of evaluating three progressively challenges:
1) \bma evaluates the robustness of skill policies against a single observation modification caused by the preceding skill. For example, as shown in Figure~\ref{fig:osm_illustration}, we evaluate the robustness of the \texttt{PlaceObject(potato, bowl)} policy under a single modification (i.e., a potato placed in the bowl). 
2) \bmb builds on this by evaluating the cumulative negative effects of multiple modifications introduced by several preceding skills, creating a more complex and challenging scenario than \bma. For instance, in the example from Figure~\ref{fig:osm_illustration}, adding a new skill \texttt{OpenDrawer(cabinet)} before \texttt{PlaceObject(potato, bowl)} results in accumulated modifications for the \texttt{MoveContainer(bowl, cabinet)} policy (i.e., the drawer is open, and a potato is placed in the bowl).
3) \bmc focuses on a realistic long-horizon task comprising a three-skill chain, testing policy performance across the entire sequence (e.g., \texttt{OpenDrawer(cabinet)} $\rightarrow$ \texttt{PlaceObject(potato, bowl)} $\rightarrow$ \texttt{MoveContainer(bowl, cabinet)}). 

We assess four popular IL algorithms using these benchmarks, offering detailed analyses of the results to generate insights and set baselines for future research (Section~\ref{sec:bm_results}). 
In addition, we examine whether data augmentation can mitigate the \pb problem by expanding the existing Libero dataset with our Rule-based Automatic Modification Generator (RAMG). This generator produces a large, visually diverse dataset, serving as a valuable resource for the community. However, comprehensive experiments on the augmented data reveal that data augmentation alone is insufficient to address the \pb, providing key insights for future research (Section~\ref{sec:da}). In summary, our contributions of this work are three-fold:



\begin{enumerate}
    
    \item For the first time, we formulate the \textbf{Observation Space Shift (\pb)}, a critical problem in long-horizon robotic tasks.
    
    \item We introduce \bm, a comprehensive benchmark that evaluates four IL methods across three increasingly challenging scenarios of \pb in long-horizon manipulation.
    
    \item By creating a large and diverse dataset with the proposed Rule-based Automatic Modification Generator, we demonstrate that data augmentation alone is insufficient to mitigate \pb, emphasizing the need and room for algorithmic solutions in future research.
\end{enumerate}


\section{Related Work}

\subsection{Robot Learning Benchmarks}


In recent years, numerous benchmarks for robot learning research have been developed. For example, Meta-World~\cite{yu2020meta} targets meta-reinforcement learning and multi-task learning. RLBench~\cite{james2020rlbench} provides 100 simulated household tasks. D4RL~\cite{fu2020d4rl} focuses on offline reinforcement learning across diverse tasks. Robomimic~\cite{robomimic2021} emphasizes learning from human demonstrations. ManiSkill2~\cite{mu2021maniskill} supports a wide range of tasks incorporating both proprioceptive and visual data. However, none of these benchmarks are designed explicitly for long-horizon tasks.

On the other hand, CausalWorld~\cite{ahmed2020causalworld} includes long-horizon tasks but focuses on causal structures and transfer learning. Calvin~\cite{mees2022calvin} addresses language-conditioned multi-task policy learning. FurnitureBench~\cite{heo2023furniturebench} features realistic, long-horizon furniture assembly tasks. RoboCasa~\cite{nasiriany2024robocasa} offers a variety of long-horizon tasks set in kitchen environments, created using generative models. LoHoRavens~\cite{zhang2023lohoravens} is tailored for language-conditioned long-horizon tasks. Despite their significant contributions, these benchmarks do not focus on skill transitions and are thus not well-suited for studying the \pb problem, which demands diverse skill transitions within long-horizon tasks.
Libero~\cite{liu2024libero}, while primarily designed for lifelong robot learning, provides diverse tasks and customization capabilities, making it an ideal foundation for developing our \bm benchmark, which provides diverse and large-scale skill transitions, and specifically targets the \pb problem.

\subsection{Skill Chaining}

Skill chaining, a fundamental structure in HIL, sequentially executing pre-learned skill policies to complete long-horizon tasks~\cite{konidaris2009skill, garrett2021integrated, guo2023recent, ahn2022can, singh2023progprompt, du2023video}, where ensuring seamless transitions between adjacent skills is the primary challenge. One approach addresses this by learning transition policies to bridge the gap between the terminal state of the previous skill and the initial state of the next~\cite{lee2019composing, byun2021training, watahiki2022one}. Lee et al., 2019 proposed transition policies with proximity predictors to connect primitive skills for sequential skills~\cite{lee2019composing}, while adversarial inverse reinforcement learning is used to learn transition policies by matching state and action distributions~\cite{byun2021training}. Goal-conditioned policies have also been employed to imitate transitions from target demonstrations~\cite{watahiki2022one}. Another approach focuses on enforcing alignment between the terminal state set of the previous skill and the initial state set of the next. Expanding the initial state set of the next skill to include the terminal state set of the previous one is one method~\cite{clegg2018learning}, though it can lead to significant state space growth over time. Lee et al., 2021 address this issue by using an adversarial learning framework to regularize the terminal state distribution of the previous skill to match the initial state distribution of the next skill~\cite{lee2021adversarial}. Additionally, contrastive learning has been applied to learn auxiliary rewards that guide terminal states closer to the initial states of the next skill~\cite{li2024auxiliary}. Chen et al., 2023, optimized entire policy chains to ensure dynamic transition feasibility in dexterous tasks~\cite{chen2023sequential}. Despite these efforts, all methods rely on the two strong assumptions outlined in Section~\ref{sec:introduction}: the reachability of initial states for each skill and the feasibility of continuous policy fine-tuning. These assumptions significantly limit their ability to address the \pb problem effectively in general scenarios.

\section{Problem Definition}
\label{sec:problem_definition}



We focus on long-horizon tasks that chain skill policies trained through imitation learning using visual inputs, assuming a robust planner generates the skill sequence based on a high-level task goal.

We first model the environment as a Partially Observable Markov Decision Process (POMDP), defined as $\mathcal{M} = \langle \mathcal{S}, \mathcal{O}, \mathcal{A}, \mathcal{T}, \mathcal{Z}, r, \gamma \rangle$. Here, $\mathcal{S}$, $\mathcal{O}$, and $\mathcal{A}$ represent the state, observation, and action spaces, respectively. The transition and observation probabilities are $\mathcal{T}(s_{t+1} \mid s_t, a_t)$ and $\mathcal{Z}(o_t \mid s_t)$. The reward function $r(s_t, a_t, s_{t+1})$ and discount factor $\gamma$ guide the trade-off between immediate and future rewards. In our setup, the observation space $\mathcal{O}$ includes both third-person camera views and proprioceptive states. The objective is, for each skill, to learn a policy $\pi : \mathcal{O} \rightarrow \Delta(\mathcal{A})$ that replicates demonstrated behavior from observation-action pairs $\tau = (o_0, a_0, \cdots, o_t, a_t)$, without explicit access to $\mathcal{S}$.

A significant problem arises during deployment when chaining skill policies, which we define as the \textbf{Observation Space Shift (\pb)}. We adopt the framework of Task and Motion Planning (TAMP) to formalize this problem~\cite{garrett2021integrated, guo2023recent}. We define predicates $\Psi = \{\psi_1, \psi_2, \ldots, \psi_n\}$ as properties or relationships within the state space, where each predicate $\psi_i$ maps a state $s \in \mathcal{S}$ to a truth value, $\psi_i(s) \in \{0, 1\}$. Essentially, each predicate acts as a binary classifier over the state space, identifying a subset of $\mathcal{S}$ where the predicate evaluates to true. As the example shown in Figure~\ref{fig:osm_illustration}, the predicate \texttt{In(potato, bowl)} evaluates to 1 if the potato is inside the bowl, and 0 otherwise. Operator is defined as $\texttt{op} = \langle \texttt{Pre}, \texttt{Eff}, c \rangle$ that comprises three components: preconditions $\texttt{Pre} \subseteq \Psi$, effects $\texttt{Eff} \subseteq \Psi$, and a cost $c$. Intuitively, an operator specifies when a skill can execute (i.e., $\texttt{Pre}$) and what is expected after execution (i.e., $\texttt{Eff}$). Symbolic operators use preconditions (e.g., \texttt{On(bowl, table)}) and effects (e.g., \texttt{On(bowl, cabinet)}) to represent only the elements (e.g., bowl instead of potato) necessary for feasibility of a skill (e.g., \texttt{MoveContainer(bowl, cabinet)}), abstracting away irrelevant ones (e.g., \texttt{In(potato, bowl)}). Formally, a predicate $\psi_i \in \Psi$ is irrelevant for an operator $\texttt{op}$ if $\psi_i \notin \texttt{Pre} \cup \texttt{Eff}$, meaning it does not influence the feasibility or expected outcome of the skill execution. However, these skill-irrelevant elements often appear in visual observations and can be unintentionally modified by preceding skills' effects. Such changes disrupt visuomotor policies, causing failures in skill execution. This problem, where changes in irrelevant predicates within the visual observation space $\mathcal{O}$ hinder visuomotor performance, is defined as \pb.

\section{The \bm Benchmark}
\label{sec:bm_challenge}

\begin{figure*}[ht]
    \centering
    \includegraphics[width=1.0\textwidth]{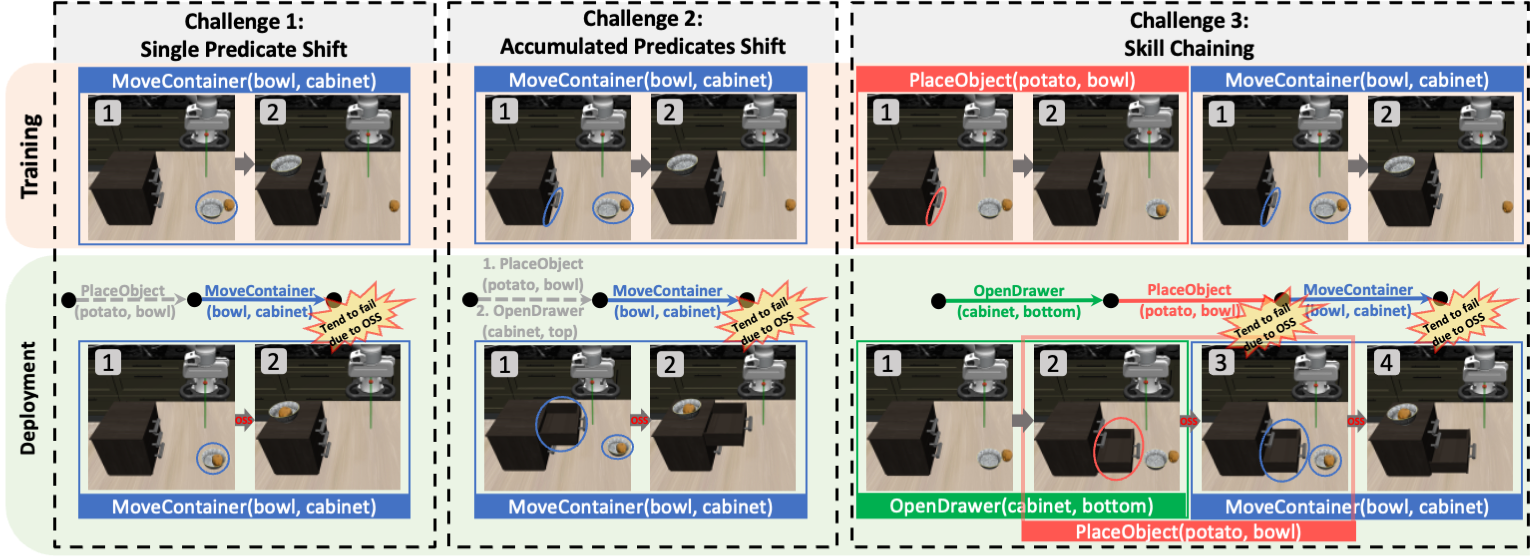} 

    \caption{This figure illustrates the three challenges of \bm, each examining a distinct aspect of \pb, using concrete examples: \textcolor{mygreen}{\texttt{OpenDrawer(cabinet, bottom)} (green)}, \textcolor{myred}{\texttt{PlaceObject(potato, bowl)} (red)}, and \textcolor{myblue}{\texttt{MoveContainer(bowl, cabinet)} (blue)}. Challenge 1, Single Predicate Shift (\bma), shows a case where \pb occurs due to the modification of a single predicate (i.e., the circle in the figure) caused by the effect of the previous skill (e.g., \texttt{IN(potato, bowl)}). Challenge 2, Accumulated Predicate Shift (\bmb), highlights the scenario where \pb arises from multiple predicate changes (i.e., circles in the figure) due to accumulated effects from preceding skills (e.g., \texttt{IN(potato, bowl)} and \texttt{DrawerOpen(cabinet, top)}). Challenge 3, Real Long-Horizon Task (\bmc), showcases how \pb impacts a real long-horizon task with three skills, where ``Single Predicate Shift'' and ``Accumulated Predicate Shift'' occur in the second skill and the third skill respectively, significantly degrading the final task performance.}
    \vspace{-1.6em}
    \label{fig:bm_arch}
\end{figure*}

We introduce the \bm benchmark, designed to facilitate a comprehensive empirical study of the \pb problem across modern IL methods. To address this issue, we build the environment using the Libero simulation platform~\cite{liu2024libero} (Section~\ref{subsec:bm_env}) and form the task set by leveraging selected existing tasks from Libero and scalably generating their modified versions (Section~\ref{subsec:bm_task}). \bm includes three challenges (Section~\ref{subsec:bm_1}~$\sim$~\ref{subsec:bm_3}) designed not only to validate the existence of \pb but also to demonstrate its significant negative impact on the success of long-horizon tasks. These challenges evaluate IL methods on the curated tasks within the constructed environment and feature 44, 88, and 10 tasks respectively, providing a diverse and extensive evaluation.


\subsection{Environment}
\label{subsec:bm_env}
We build the environment of \bm on the Libero platform~\cite{liu2024libero}, which, although originally designed for lifelong robot learning tasks, provides diverse manipulation scenes featuring a Franka Emika Panda robot arm and accommodates a wide variety of tasks. Libero is highly flexible for creating and customizing tasks because they are generated using Planning Domain Definition Language (PDDL) files, which specify the planning problem including operators, predicates, and their relationships. These features make Libero an excellent foundation for developing \bm, enabling us to adapt it specifically to study the \pb problem.

\subsection{Task Design}
\label{subsec:bm_task}

Building on the Libero environment, \bm focuses on studying the impact of \pb on long-horizon tasks. To achieve this, we require a set of atomic robotic tasks, where each task involves only a single skill. These tasks are used to simulate individual skills within a skill chain that are unaffected by \pb. Additionally, we generate modified counterparts to simulate scenarios where \pb occurs. This setup enables performance comparisons between the two sets when evaluating IL methods. All three challenges are based on these 2 sets of tasks. 

To build the set of tasks unaffected by \pb, we select all the skill-level tasks from Libero-100, the most comprehensive and diverse task suite in Libero, spanning 12 manipulation scenes and covering a wide range of object interactions and motor skills. Multi-skill tasks (e.g., “open the top drawer of the cabinet and put the bowl in it”) are excluded, while single-skill tasks (e.g., “open the bottom drawer of the cabinet”) are retained. As a result, the final set comprises 44 single-skill tasks. 


To build the set of counterparts, we propose the \textbf{Rule-based Automatic Modification Generator (RAMG)}, which can scalably generate modified tasks. RAMG is an algorithm designed to enhance the visual diversity of skills in predefined environments through structured, rule-based modifications. Operating on PDDL files, RAMG iteratively modifies predicates, $\Psi$, by altering object positions, introducing new objects, or changing object states within the environment. It supports three types of modifications: (1) repositioning existing objects or adding external objects to specific regions, (2) changing the state of fixtures (e.g., toggling a predicate like \texttt{DrawerOpen(bottom\_drawer, cabinet)}), and (3) modifying containment relations by altering predicates like \texttt{In(potato, bowl)}. These modifications adhere to dynamic constraints (e.g., newly added objects do not interfere with existing \texttt{Pre} and \texttt{Eff} conditions) and maintain logical consistency (e.g., preventing contradictions where a predicate $\psi_i$ and its negation $\neg\psi_i$ hold simultaneously). Using deterministic yet flexible rules, RAMG provides a scalable method for generating task variations. With a single modification per task, RAMG can produce up to 1,727 modified tasks from the 44 selected tasks, and adding multiple modifications significantly increases this number. The extensive set of generated modified tasks offers flexibility and a robust foundation for designing the three challenges.


\subsection{Challenge 1: Single Predicate Shift (\bma)}
\label{subsec:bm_1}



As outlined in Section~\ref{sec:problem_definition}, \pb arises when skill-irrelevant predicates, $\Psi$, are altered in ways that do not affect the feasibility of the current skill but disrupt the execution of its visuomotor policy. For example, as shown in the left box of Figure~\ref{fig:bm_arch}, the predicate \texttt{In(potato, bowl)} does not impact the feasibility of the current skill, \texttt{MoveContainer(bowl, cabinet)}, since it is absent from $\texttt{Pre}$ and $\texttt{Eff}$. However, changes to this predicate can still alter the visuomotor observation space, $\mathcal{O}$, negatively affecting the skill policy's performance. In the context of skill chaining, a foundational structure in long-horizon tasks, the preceding skill is the most likely to introduce modifications in $\mathcal{O}$ that induce \pb for the subsequent skill. Therefore, we introduce a challenge, Single Predicate Shift (\bma), specifically designed to evaluate the impact of \pb caused by single-step transitions on the performance of baseline methods (see examples in Figure~\ref{fig:bm_arch}).

To investigate the impact of single-step \pb on baseline methods, we compare their performance on skills unaffected by \pb (e.g., \texttt{On(potato, table)}) versus those affected by \pb (e.g., \texttt{In(potato, bowl)}). As detailed in Section~\ref{subsec:bm_task}, for the former, we evaluate baselines on the 44 selected tasks. For the latter, we use RAMG to generate 44 corresponding randomly modified tasks, each with a single modification applied to simulate the occurrence of \pb caused by the preceding skill. 


\subsection{Challenge 2: Accumulated Predicate Shift (\bmb)}
\label{subsec:bm_2}

The effect of executing a skill persists until a future skill reverses it, meaning it impacts not only the immediate next skill but also multiple subsequent skills. Over the course of a skill chain, effects from numerous skills can accumulate, resulting in a more severe \pb for a future skill. Formally, given a sequence of operators $\{\texttt{op}_1, \texttt{op}_2, \dots, \texttt{op}_t\}$, the accumulated effects at time step $t$ are given by $\bigcup_{i=1}^{t} \texttt{Eff}_i$. As illustrated in the middle box of Figure~\ref{fig:bm_arch}, the accumulated effects, \texttt{On(plate, cabinet)} and \texttt{In(potato, bowl)}, collectively impact the current skill, \texttt{MoveContainer(bowl, cabinet)}, by modifying predicates in the observation space $\mathcal{O}$. Thus, it is valuable to study the differences between accumulated \pb and single-step \pb. To address this, we introduce another challenge, Accumulated Predicate Shift (\bmb), specifically designed to analyze the impact of \pb accumulation across preceding skills (check examples in Figure~\ref{fig:bm_arch}).

Similar to \bma, we evaluate baseline performance on the next skill under two conditions: task unaffected by \pb and task modified by previous effects. Using PDDL description files, we create two sets of modified tasks, each simulating the cumulative effects of multiple preceding skills on the current skill, with one set incorporating two modifications and the other three. However, multiple modifications can sometimes conflict. For example, a modification such as ``place an apple inside the small bowl'' could conflict with another that specifies ``place a potato inside the small bowl,'' potentially causing the Libero environment to break due to overlapping space constraints. By iteratively invoking the proposed RAMG two or three times, we easily generate the two sets of modified tasks, leveraging RAMG's ability to ensure constraint-compliant task generation.


\subsection{Challenge 3: Skill Chaining (\bmc)}
\label{subsec:bm_3}
The ultimate goal of addressing the \pb problem is to improve the success of long-horizon robot tasks. To this end, we introduce a challenge, Skill Chaining (\bmc), consisting of 10 long-horizon tasks, each comprising a chain of three skills (check examples in Figure~\ref{fig:bm_arch}). This challenge serves as a straightforward way to demonstrate the impact of the \pb problem on long-horizon task performance.

To construct this challenge, we manually select and combine skills from the 44 selected skill-level tasks, ensuring that \pb occurs in each skill while avoiding conflicts between modifications. We reset the robot to a neutral position after each skill to eliminate dynamic transition feasibility issues~\cite{chen2023sequential}, ensuring the challenge focuses solely on the impact of \pb, which specifically addresses the negative effects of changes in visual observations.

\section{The \bm Experimental Results}
\label{sec:bm_results}

\subsection{Experimental Setup}

\subsubsection{Baselines}
\label{subsec:basic_bl}

We select four widely used imitation learning algorithms, representing diverse architectures and design approaches for baseline comparisons, to learn each skill and evaluate the impact of \pb on their performance in long-horizon tasks. Among these, three are Behavioral Cloning approaches from Libero, while one is a vision-language-action model.

\textbf{Behavioral Cloning (BC) in Libero}: A set of three BC algorithms~\cite{torabi2018behavioral} from Libero is adopted: BC-RESNET-RNN~\cite{mandlekar2021matters}, BC-RESNET-T~\cite{zhu2023viola}, and BC-VIT-T~\cite{kim2021vilt}. These algorithms feature diverse neural network architectures for visual and language encoding. All three use BERT embeddings~\cite{devlin2018bert} to encode the language instructions for each skill. In BC-RESNET-RNN, ResNet~\cite{he2016identity} serves as the visual backbone for encoding per-step visual observations, while an LSTM processes the sequence of encoded visual embeddings as the temporal backbone. BC-RESNET-T employs the same visual encoder, ResNet, but replaces the LSTM with a transformer decoder~\cite{vaswani2017attention} for temporal processing. BC-VIT-T uses Vision Transformer (ViT)~\cite{dosovitskiy2020image} as the visual backbone and a transformer decoder as the temporal backbone. All these BC algorithms output a multi-modal distribution over manipulation actions using a Gaussian Mixture Model (GMM) output head~\cite{bishop1994mixture}, from which an action is sampled.


\textbf{OpenVLA}: Applying large foundation models, such as Large Language Models (LLMs)~\cite{openai2024chatgpt} and Vision-Language Models (VLMs)~\cite{dubey2024llama}, to robotics has gained popularity, leading to vision-language-action (VLA) models. Among these, OpenVLA~\cite{kim2024openvla}, an open-sourced VLA model, outperforms other state-of-the-art methods, making it a natural choice for evaluating the \pb problem as a representative generalist policy. The language description, conditioned on each task and sourced from Libero, remains consistent, whether or not the task is affected by \pb.

\subsubsection{Metrics}
We mainly use two metrics in all the experiments: 

\textbf{Ratio Performance Delta (RPD)}: We define the metric ``Ratio Performance Delta'' as the relative change in skill success rate caused by \pb. It is calculated as the difference between a baseline's success rate on the original skill-level task and its success rate on the modified task where \pb occurs, normalized by the original success rate. A positive RPD indicates that \pb negatively impacts the current skill, while an RPD less than or equal to zero suggests no negative effect. Instances of negative RPD can occur due to the inherent randomness of IL models.

\textbf{Delta to Upper Bound Ratio (DUBR)}: Unlike \bma and \bmb, which evaluate the performance delta for individual skills, \bmc assesses the performance delta across an entire skill chain. To support this, we introduce a new metric. First, we define the ``Chain Upper Bound'' as the product of success rates for each skill in the chain when evaluated without \pb occurrence, representing the maximum achievable success rate in the absence of \pb. Then, we introduce the ``Delta to Upper Bound Ratio'', calculated as the difference between the actual success rate of completing the entire skill chain and the ``Chain Upper Bound'', divided by the ``Chain Upper Bound'', providing a normalized measure of the performance delta for the entire chain.

\subsubsection{Implementation Details}
For observation space design, we align with the baselines' default setups. OpenVLA uses only a third-person camera view as its observation space. To maintain consistency in visual observations, we define the observation space for BCs as a combination of third-person camera images, 7-DoF robot arm joint angles, and 2-DoF parallel gripper joint states, excluding only the wrist-camera view. For all the baselines, the action space is defined as a 7-dimensional relative Cartesian displacement (w.r.t. the gripper frame), and the control frequency is set to 20 Hz. For all the experiments, we use three random seeds and report only the averaged results across these runs.

\subsection{Results for \bma}
\label{subsec:results_bma}

\begin{figure*}[ht]
    \centering
    \includegraphics[width=1.0\textwidth]{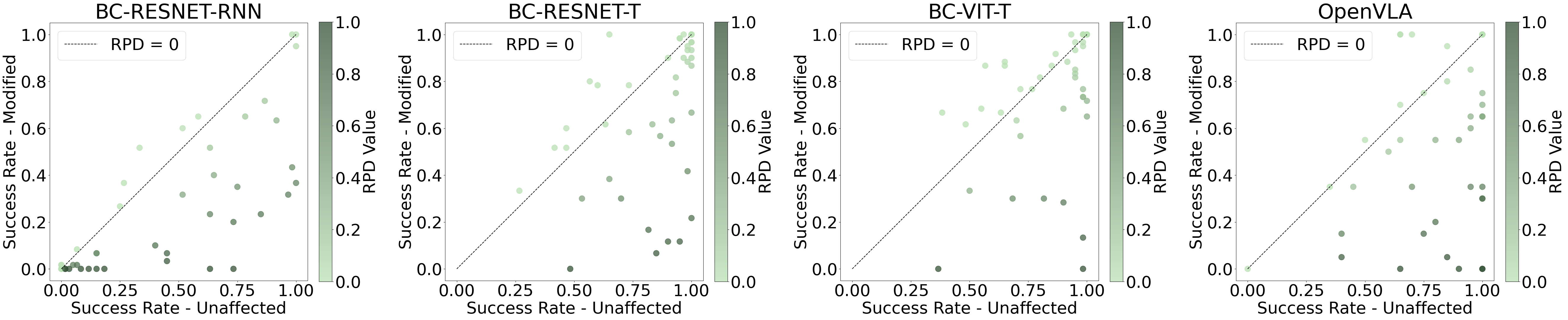} 
    \caption{This figure presents the results for \bma. In each baseline subfigure, the majority of points lie below the diagonal line, representing tasks with a positive Ratio Performance Delta, indicating that single predicate modification negatively affects tasks performance.}
    \label{fig:ch1_results}
    \vspace{-1.5em}
\end{figure*}

We evaluate baseline methods, including BC-RESNET-RNN, BC-RESNET-T, BC-VIT-T, and OpenVLA, on the 44 selected tasks and their 44 modified counterparts, calculating the Ratio Performance Delta for each baseline. The results are presented in Figure~\ref{fig:ch1_results}, where the x-axis represents the success rate on tasks unaffected by \pb, and the y-axis represents the success rate on their modified counterparts. Each point corresponds to a task pair (unaffected and modified), with darker colors indicating higher Ratio Performance Delta values. A diagonal line separates tasks: points on or above the line have non-positive Ratio Performance Delta, indicating no negative impact from \pb, while points below the line signify tasks negatively affected by \pb. As shown in Figure~\ref{fig:ch1_results}, the percentage of tasks negatively affected by \pb (points below the diagonal) is 68\%, 66\%, 50\%, and 66\% for BC-RESNET-RNN, BC-RESNET-T, BC-VIT-T, and OpenVLA, respectively. Among these tasks, the average Ratio Performance Delta is 67\%, 35\%, 34\%, and 54\% for the respective baselines, highlighting the substantial performance degradation caused by \pb. These results demonstrate that \pb frequently occurs even in single-step transitions, emphasizing its potential to jeopardize the success of long-horizon tasks.


\subsection{Results for \bmb}
\label{subsec:results_bmb}
\vspace{-1.2em}


\begin{figure}[ht]
    \centering
    \includegraphics[width=0.5\textwidth]{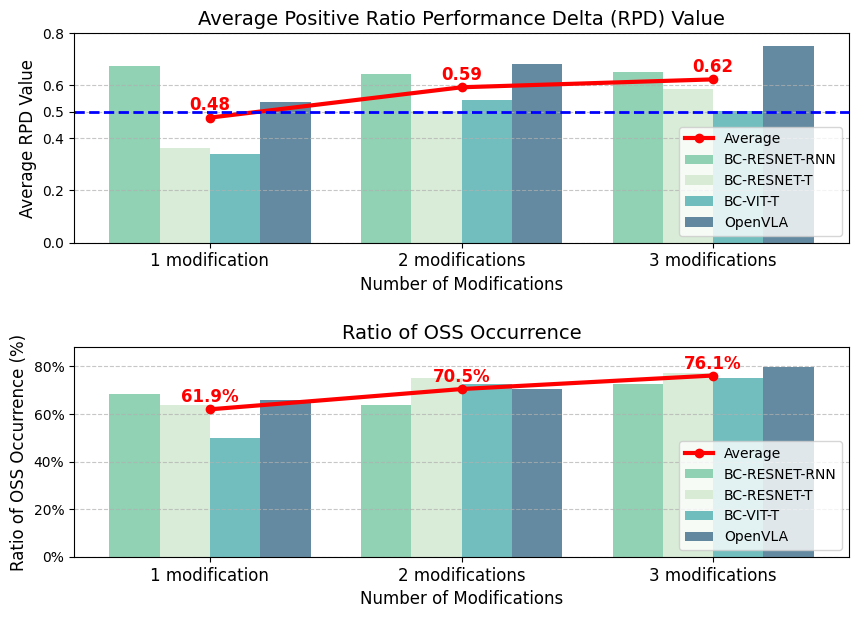} 
    \caption{This figure shows the results for \bmb. Two bar charts summarize: (top) the average positive Ratio Performance Delta for sets with different numbers of modifications, and (bottom) the average ratio of \pb occurrence across sets with varying numbers of modifications. The upward trend of the red lines in both bar charts indicates that the accumulation of \pb progressively exacerbates its negative impact on long-horizon task completion, both in magnitude and frequency.}
    \vspace{-0.5em}
    \label{fig:ch2_results}
\end{figure}

As detailed in Section~\ref{subsec:bm_2}, we use the proposed RAMG to generate two sets of modified tasks: one with two modifications and another with three, simulating scenarios of accumulated modifications. Similar to \bma, we evaluate baselines on the 44 unaffected tasks and their corresponding modified counterparts. Figure~\ref{fig:ch2_results} presents the evaluation results for \bmb alongside \bma (i.e., single modification), yielding three sets of results for comparison. Two bar charts summarize key findings: (top) the average positive Ratio Performance Delta across the three sets and (bottom) the average occurrence ratio of \pb (i.e., cases where Ratio Performance Delta is positive). Each bar chart consists of three grouped bars, where each group corresponds to sets with one, two, or three modifications, and within each group, bars represent different baselines. Additionally, a red line denotes the average performance across all baselines for each set. In both charts, this red line follows an increasing trend, indicating that as the number of modifications grows, the negative impact on task performance intensifies, both in magnitude and frequency. Notably, across all baselines, the average Ratio Performance Delta for sets with multiple modifications exceeds 50\% for all baselines (blue line in the top bar chart), signifying a substantial performance decline that is likely to cause failures in long-horizon tasks. Given that skill effect accumulation is inherent in long-horizon tasks, with modifications compounding along the skill chain, these results highlight that \pb presents a significant obstacle to achieving reliable performance.


\subsection{Results for \bmc}
\label{subsec:results_bmc}

\begin{figure*}[ht]
    \centering
    \includegraphics[width=1.0\textwidth]{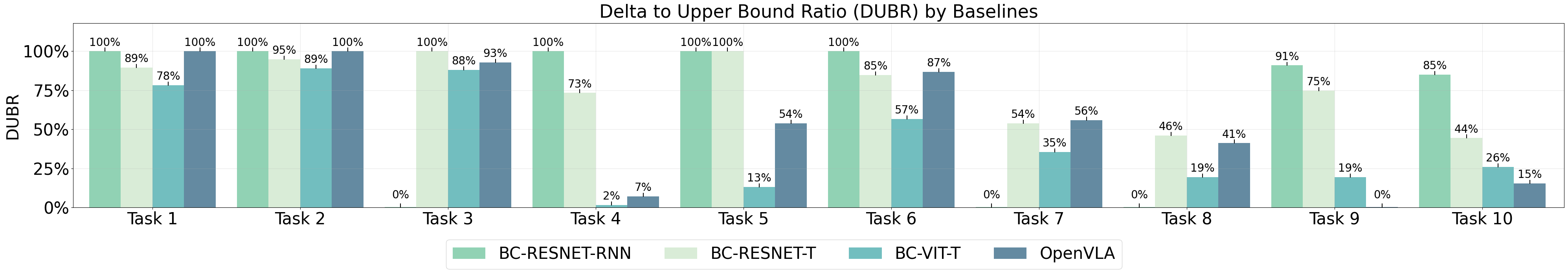} 
    \caption{This figure presents the results for \bmc, where the ``Delta to Upper Bound Ratio'' (bars) is positive and notably high in most cases, highlighting the substantial negative impact of \pb on long-horizon task completion. 
    }
    \vspace{-1.75em}
    \label{fig:ch3_results}
\end{figure*}

We include \bmc to evaluate the impact of \pb on skill chaining. As described in Section~\ref{subsec:bm_3}, we test baselines on 10 manually designed tasks, each comprising a skill chain of three skills, and use the ``Delta to Upper Bound Ratio'' metric to quantify \pb's effect on performance. Figure~\ref{fig:ch3_results} presents the results, the ``Delta to Upper Bound Ratio'' is shown as bar charts, where different baselines are shown in varied colors. The ``Delta to Upper Bound Ratio'' values are positive and high in most cases. Note that some bars for BC-RESNET-RNN show 0\% values because the ``Chain Upper Bound'' for those tasks is already 0\%, indicating that BC-RESNET-RNN fails even without \pb. These results further emphasize \pb's detrimental effect on long-horizon task completion.
\section{Can Data Augmentation Mitigate \pb?}
\label{sec:da}

\subsection{Data Augmentation}
\label{subsec:our_bl}

As defined in Section~\ref{sec:problem_definition}, the \pb problem occurs when visual changes caused by preceding skills disrupt the execution of the current skill. A straightforward solution is to collect demonstrations with more diverse visual content, allowing algorithms to learn from varied observation spaces and become more robust to novel visual modifications. To simulate this approach, we augment the demonstrations and train all the baselines on an expanded dataset.

Using RAMG, we generate diverse modified tasks, creating a total of 1,727 new tasks from the 44 selected tasks with a single modification applied to each. For each modified task, we replay demonstrations in the corresponding environment, filtering out failed replays to produce new demonstrations where the trajectory remains unchanged, yet with varied visual observations. This method produces a significantly larger dataset, totaling 57,000 demonstrations—nearly 30 times the size of the original Libero dataset for the 44 selected tasks (2,000 demonstrations). Notably, this figure corresponds to cases with a single modification applied; an even much larger dataset can be generated by iteratively using RAMG to apply multiple modifications. We then pre-train (BCs) or fine-tune (OpenVLA) all baselines using this expanded dataset.


\begin{table}[ht]
\centering
\caption{Comparison of baseline performance on Setup A and Setup B. Setup A: baselines trained on original demonstrations and evaluated on skills affected by \pb. Setup B: baselines trained on augmented demonstrations and evaluated on the same skills affected by \pb. The comparison reflects the difference in results between the two setups.}
\label{tb:da_results}
\begin{tabular}{c|c|c|c}
\toprule
                        & Setup A        & Setup B        & \begin{tabular}[c]{@{}c@{}}Comparison \\ (A - B)\end{tabular} \\ \midrule
BC-RESNET-RNN           & 0.25          & 0.31           & -0.06                                                      \\ 
BC-RESNET-T             & 0.67          & 0.54           & 0.13                                                       \\ 
BC-VIT-T                & 0.74          & 0.64           & 0.10                                                       \\ 
OpenVLA                 & 0.54          & 0.54           & 0.00                                                       \\ \bottomrule
\end{tabular}
\vspace{-2em}
\end{table}

\subsection{Results}
\label{subsec:results_bl}


To investigate whether the data augmentation method mitigates the \pb problem, we compare performance between two setups to assess whether improvements exist on skills affected by \pb. Setup A: baselines are trained on original demonstrations for a skill and evaluated on the skill affected by \pb. Setup B: baselines are trained on augmented demonstrations and evaluated on the same skill affected by \pb. Table~\ref{tb:da_results} presents the performance values for both setups across all baselines, with averaged results reported for all 44 tasks. A comparison between the setups reveals that BC-RESNET-RNN shows a slight improvement under Setup B. However, other baselines, including BC-RESNET-T, BC-VIT-T, and OpenVLA, perform equivalently or worse, suggesting that data augmentation alone is limited in mitigating \pb. While larger and more diverse datasets generally improve performance, they may not fully address \pb. One possible reason is that some baselines struggle to generalize across the diverse visual distributions of demonstrations, leading to even worse performance. Another challenge is the combinatorial explosion of scene and task variations, making it impractical to cover all relevant cases through data augmentation alone. This limitation persists even with RAMG, which is constrained to simulated data generation and cannot scale to real-world data. These findings underscore the need for tailored algorithmic solutions to effectively address \pb.


\section{Conclusion}

This work introduces and investigates Observation Space Shift (\pb), a critical issue that hinders the completion of long-horizon robot tasks. Through three challenges in \bm, we validate the impact of \pb on several the performance of several popular baseline algorithms and present key findings:

\begin{itemize}
    \item \textbf{\pb is a common and severe problem}: Results on \bm demonstrate that \pb can substantially degrade task completion, especially in long-horizon tasks where visual modifications accumulate over time.
    \item \textbf{Visual modification magnitude matters, task difficulty may not}: Larger visual modifications lead to greater \pb severity, while task difficulty was not significantly correlated, making \pb particularly problematic as task horizons grow.
    \item \textbf{Data augmentation may not be sufficient}: While increasing dataset size and diversity generally improves performance, we found that it did not fully mitigate \pb and may even degrade performance in some cases. A key challenge is that data augmentation alone cannot generate a dataset large and diverse enough to cover the vast combinatorial space of visually varying scene setups, which arise from the numerous skill compositions within long-horizon tasks. Furthermore, current baseline architectures are not explicitly designed to address \pb, highlighting the need for tailored algorithmic solutions, such as mechanisms that focus on task-relevant visual cues, and reduce sensitivity to irrelevant variations.
\end{itemize}

These findings highlight that effectively addressing \pb requires either more advanced methods for scaling the visual diversity of robotic data or innovative algorithmic designs that enhance robustness in long-horizon tasks. We hope our benchmarks and insights will inspire further research toward tackling this critical challenge.


\bibliographystyle{IEEEtran}
\bibliography{refs}

\section{Appendix}

\subsection{Factors Influencing \pb}
\label{sec:factors}


In this section, we investigate two potential factors influencing the severity of \pb: skill difficulty and the magnitude of visual modifications. This analysis aims to provide insights for future research and inform the development of algorithms robust to \pb. We define the following two hypotheses:

\begin{enumerate}[I:]
    \item Higher skill difficulty (i.e., lower skill success rate) leads to more severe performance degradation caused by \pb for that skill.
    \item Larger magnitudes of visual modifications result in more severe performance degradation caused by \pb for the current skill.
\end{enumerate}

To test these hypotheses, we use Spearman's rank correlation coefficient ($\rho$)~\cite{spearman1961proof}, which effectively measures the strength of monotonic relationships between two variables. If hypothesis I holds, we expect a significant negative monotonic relationship (i.e., $\rho < 0$, $p < 0.05$) between skill success rate and the severity of \pb. Similarly, if hypothesis II is valid, we anticipate a significant positive monotonic relationship (i.e., $\rho > 0$, $p < 0.05$) between the magnitude of visual modifications and the severity of \pb.

\subsubsection{Test Hypothesis I (Factor: Task Difficulty)}
\begin{table}[ht]
\centering
\caption{Spearman’s rank correlation coefficients ($\rho$) and p-values ($p$) for the relationship between task difficulty and OSS severity.}
\label{tb:fac1_results}
\begin{tabular}{|c|c|c|c|}
\hline
      & BC-RESNET-RNN                                                & BC-RESNET-T                                                  & BC-VIT-T                                                     \\ \hline
Set 1 & \begin{tabular}[c]{@{}c@{}}$\rho$: -0.86\\ $p$: 0.06\end{tabular} & \begin{tabular}[c]{@{}c@{}}$\rho$: 0.02\\ $p$: 0.97\end{tabular}  & \begin{tabular}[c]{@{}c@{}}$\rho$: 0.04\\ $p$: 0.94\end{tabular}  \\ \hline
Set 2 & \begin{tabular}[c]{@{}c@{}}$\rho$: 0.89\\ $p$: 0.04\end{tabular}  & \begin{tabular}[c]{@{}c@{}}$\rho$: -0.04\\ $p$: 0.94\end{tabular} & \begin{tabular}[c]{@{}c@{}}$\rho$: 0.78\\ $p$: 0.04\end{tabular}  \\ \hline
Set 3 & \begin{tabular}[c]{@{}c@{}}$\rho$: -0.80\\ $p$: 0.20\end{tabular} & \begin{tabular}[c]{@{}c@{}}$\rho$: 0.50\\ $p$: 0.39\end{tabular}  & \begin{tabular}[c]{@{}c@{}}$\rho$: 0.55\\ $p$: 0.33\end{tabular}  \\ \hline
Set 4 & \begin{tabular}[c]{@{}c@{}}$\rho$: 0.95\\ $p$: 0.05\end{tabular}  & \begin{tabular}[c]{@{}c@{}}$\rho$: -0.21\\ $p$: 0.74\end{tabular} & \begin{tabular}[c]{@{}c@{}}$\rho$: 0.15\\ $p$: 0.80\end{tabular}  \\ \hline
Set 5 & \begin{tabular}[c]{@{}c@{}}$\rho$: 0.21\\ $p$: 0.79\end{tabular}  & \begin{tabular}[c]{@{}c@{}}$\rho$: -0.34\\ $p$: 0.57\end{tabular} & \begin{tabular}[c]{@{}c@{}}$\rho$: -0.67\\ $p$: 0.22\end{tabular} \\ \hline
Set 6 & \begin{tabular}[c]{@{}c@{}}$\rho$: 0.89\\ $p$: 0.11\end{tabular}  & \begin{tabular}[c]{@{}c@{}}$\rho$: 0.79\\ $p$: 0.11\end{tabular}  & \begin{tabular}[c]{@{}c@{}}$\rho$: 0.87\\ $p$: 0.05\end{tabular}  \\ \hline
Avg.  & \begin{tabular}[c]{@{}c@{}}$\rho$: 0.21\\ $p$: 0.21\end{tabular}  & \begin{tabular}[c]{@{}c@{}}$\rho$: 0.12\\ $p$: 0.62\end{tabular}  & \begin{tabular}[c]{@{}c@{}}$\rho$: 0.29\\ $p$: 0.40\end{tabular}  \\ \hline
\end{tabular}
\end{table}

We use the metric ``Ratio Performance Delta'' to quantify the severity of \pb and measure task difficulty using task success rate, assuming demonstrations for all skills are of similar quality. To analyze the relationship between skill difficulty and \pb, we design an experiment that isolates skill difficulty as the only variable. We select six sets of skills, where all skills within each set are performed in the same scene (e.g., kitchen-scene-1 in Libero). Identical modifications are applied to all skills within each set to simulate \pb, resulting in several pairs of skills: one unaffected by \pb and the other affected by \pb. Between pairs, the only difference is the skill's difficulty. After evaluating BCs on these pairs, we generate two lists of results for each set and algorithm: one for success rates of skills unaffected by \pb (i.e., task difficulty) and another for ``Ratio Performance Delta'' on skills affected by \pb (i.e., severity of \pb). These results are used to calculate Spearman's rank correlation coefficient ($\rho$) and the corresponding p-value. Table~\ref{tb:fac1_results} presents the findings. None of the baselines show a significant negative monotonic relationship (i.e., $\rho < 0$, $p < 0.05$), indicating no evidence to support the hypothesis that task difficulty is a key factor influencing the severity of \pb.

\subsubsection{Test Hypothesis II (Factor: Magnitude of Visual Modification)}

We also use the metric ``Ratio Performance Delta'' to quantify the severity of \pb. To ensure that the magnitude of visual modification is the only variable changing, we adopt an approach similar to \bmb by increasing the number of modifications applied to the same skill. However, unlike \bmb, where modifications are generated randomly by RAMG, it is possible for skills with a larger number of modifications (e.g., affecting several small areas) to have a smaller visual modification magnitude than skills with fewer but more significant modifications (e.g., affecting a large area). To address this, we incrementally add new modifications based on existing ones, ensuring a clear correlation between the number of modifications and the size of the visual changes. After evaluating BCs on skills unaffected by \pb and skills with multiple modifications, we generate two lists for each algorithm: one representing the number of modifications (i.e., visual modification magnitude) and the other capturing the RPD (i.e., severity of \pb). Spearman's rank correlation coefficient ($\rho$) and the corresponding p-value are calculated to analyze the relationship between these two variables. For all BCs, we obtain results as $\rho = 1.0$ and $p = 0.0$, indicating a significant positive monotonic relationship between visual modification magnitude and the severity of \pb. These findings confirm that the magnitude of visual modifications is a key factor influencing the severity of \pb.

\subsection{Numerical Results for BOSS-C1 and BOSS-C2}
For better visual clarity and due to space constraints, we omit the detailed numerical results for both BOSS-C1 and BOSS-C2. Instead, this section presents tables summarizing the quantitative results for both BOSS-C1 and BOSS-C2.

\subsubsection{Numerical Results for BOSS-C1}
        
        \begin{table*}[]
        \scriptsize
        \setlength{\tabcolsep}{2pt} 
        \renewcommand{\arraystretch}{1.1} 
        \centering
        \caption{BOSS-C1 Results: Evaluation of 4 baselines on the 44 selected tasks and their counterparts, highlighting positive RPD results.}
        \label{tb:bm1_results_appendix}
        \begin{tabular}{|l|l|l|l|l|l|l|l|l|l|l|l|l|l|l|l|l|l|l|l|l|l|l|l|}
        \hline
                                   &          & 0 & 1 & 2 & 3 & 4 & 5 & 6 & 7 & 8 & 9 & 10 & 11 & 12 & 13 & 14 & 15 & 16 & 17 & 18 & 19 & 20 & 21 \\ \hline
        \multirow{3}{*}{BC-RESNET-RNN} & SR - ORI & 0.98 & 0.73 & 0.78 & 0.97 & 0.00 & 0.02 & 0.63 & 0.05 & 0.15 & 0.00 & 0.07 & 0.63 & 0.00 & 0.52 & 0.40 & 1.00 & 0.98 & 0.00 & 0.27 & 0.25 & 0.63 & 0.52 \\ \cline{2-24}
                                    & SR - MOD & 1.00 & 0.20 & 0.65 & 0.32 & 0.02 & 0.00 & 0.00 & 0.02 & 0.07 & 0.00 & 0.08 & 0.23 & 0.00 & 0.60 & 0.10 & 0.37 & 0.43 & 0.00 & 0.37 & 0.27 & 0.52 & 0.32 \\ \cline{2-24}
                                    & RPD    & -0.02 & \textbf{\textcolor{red}{0.73}} & \textbf{\textcolor{red}{0.17}} & \textbf{\textcolor{red}{0.67}} & 0.00 & \textbf{\textcolor{red}{1.00}} & \textbf{\textcolor{red}{1.00}} & \textbf{\textcolor{red}{0.67}} & \textbf{\textcolor{red}{0.56}} & 0.00 & -0.25 & \textbf{\textcolor{red}{0.63}} & 0.00 & -0.16 & \textbf{\textcolor{red}{0.75}} & \textbf{\textcolor{red}{0.63}} & \textbf{\textcolor{red}{0.56}} & 0.00 & -0.37 & -0.07 & \textbf{\textcolor{red}{0.18}} & \textbf{\textcolor{red}{0.39}} \\ \hline
\multirow{3}{*}{BC-RESNET-T} & SR - ORI & 1.00 & 1.00 & 1.00 & 0.95 & 0.93 & 1.00 & 0.95 & 0.83 & 0.82 & 0.63 & 0.97 & 0.92 & 0.42 & 1.00 & 0.87 & 1.00 & 0.98 & 0.90 & 0.92 & 0.73 & 0.73 & 1.00 \\ \cline{2-24}
                                    & SR - MOD & 1.00 & 0.67 & 0.93 & 0.12 & 0.82 & 0.22 & 0.98 & 0.62 & 0.17 & 0.62 & 1.00 & 0.53 & 0.52 & 0.87 & 0.57 & 1.00 & 0.93 & 0.12 & 0.63 & 0.78 & 0.58 & 0.97 \\ \cline{2-24}
                                    & RPD    & 0.00 & \textbf{\textcolor{red}{0.33}} & \textbf{\textcolor{red}{0.07}} & \textbf{\textcolor{red}{0.88}} & \textbf{\textcolor{red}{0.13}} & \textbf{\textcolor{red}{0.78}} & -0.04 & \textbf{\textcolor{red}{0.26}} & \textbf{\textcolor{red}{0.80}} & \textbf{\textcolor{red}{0.03}} & -0.03 & \textbf{\textcolor{red}{0.42}} & -0.24 & \textbf{\textcolor{red}{0.13}} & \textbf{\textcolor{red}{0.35}} & 0.00 & \textbf{\textcolor{red}{0.05}} & \textbf{\textcolor{red}{0.87}} & \textbf{\textcolor{red}{0.31}} & -0.07 & \textbf{\textcolor{red}{0.20}} & \textbf{\textcolor{red}{0.03}} \\ \hline
\multirow{3}{*}{BC-VIT-T} & SR - ORI & 1.00 & 1.00 & 0.98 & 0.95 & 0.95 & 0.98 & 0.98 & 0.87 & 0.68 & 0.55 & 0.98 & 0.90 & 0.92 & 0.90 & 0.98 & 1.00 & 1.00 & 0.98 & 0.98 & 0.70 & 0.63 & 1.00 \\ \cline{2-24}
                                    & SR - MOD & 1.00 & 0.65 & 1.00 & 0.97 & 0.82 & 0.13 & 0.97 & 0.92 & 0.30 & 0.68 & 0.73 & 0.28 & 0.88 & 0.68 & 0.77 & 1.00 & 1.00 & 0.00 & 0.73 & 0.63 & 0.67 & 1.00 \\ \cline{2-24}
                                    & RPD    & 0.00 & \textbf{\textcolor{red}{0.35}} & -0.02 & -0.02 & \textbf{\textcolor{red}{0.14}} & \textbf{\textcolor{red}{0.86}} & \textbf{\textcolor{red}{0.02}} & -0.06 & \textbf{\textcolor{red}{0.56}} & -0.24 & \textbf{\textcolor{red}{0.25}} & \textbf{\textcolor{red}{0.69}} & \textbf{\textcolor{red}{0.04}} & \textbf{\textcolor{red}{0.24}} & \textbf{\textcolor{red}{0.22}} & 0.00 & 0.00 & \textbf{\textcolor{red}{1.00}} & \textbf{\textcolor{red}{0.25}} & \textbf{\textcolor{red}{0.10}} & -0.05 & 0.00 \\ \hline
\multirow{3}{*}{OpenVLA} & SR - ORI & 1.00 & 1.00 & 0.65 & 1.00 & 1.00 & 1.00 & 1.00 & 1.00 & 1.00 & 1.00 & 0.70 & 0.65 & 0.65 & 0.50 & 0.75 & 0.80 & 1.00 & 0.85 & 0.80 & 0.95 & 0.40 & 1.00 \\ \cline{2-24}
                                    & SR - MOD & 0.35 & 1.00 & 0.55 & 0.00 & 1.00 & 0.70 & 0.00 & 1.00 & 0.30 & 0.65 & 0.35 & 1.00 & 1.00 & 0.55 & 0.75 & 0.20 & 0.75 & 0.05 & 0.55 & 0.35 & 0.15 & 1.00 \\ \cline{2-24}
                                    & RPD    & \textbf{\textcolor{red}{0.65}} & 0.00 & \textbf{\textcolor{red}{0.15}} & \textbf{\textcolor{red}{1.00}} & 0.00 & \textbf{\textcolor{red}{0.30}} & \textbf{\textcolor{red}{1.00}} & 0.00 & \textbf{\textcolor{red}{0.70}} & \textbf{\textcolor{red}{0.35}} & \textbf{\textcolor{red}{0.50}} & -0.54 & -0.54 & -0.10 & 0.00 & \textbf{\textcolor{red}{0.75}} & \textbf{\textcolor{red}{0.25}} & \textbf{\textcolor{red}{0.94}} & \textbf{\textcolor{red}{0.31}} & \textbf{\textcolor{red}{0.63}} & \textbf{\textcolor{red}{0.62}} & 0.00 \\ \hline
\multirow{3}{*}{MaIL} & SR - ORI & 0.93 & 1.00 & 0.43 & 0.93 & 0.93 & 0.97 & 0.95 & 0.70 & 0.73 & 0.90 & 0.52 & 0.52 & 0.28 & 0.95 & 0.35 & 1.00 & 0.97 & 0.92 & 1.00 & 0.20 & 0.18 & 0.98 \\ \cline{2-24}
                                    & SR - MOD & 0.95 & 0.00 & 0.58 & 0.00 & 0.00 & 0.15 & 0.00 & 0.57 & 0.28 & 0.83 & 0.92 & 0.42 & 0.13 & 0.60 & 0.00 & 0.45 & 0.83 & 0.00 & 0.92 & 0.27 & 0.37 & 0.93 \\ \cline{2-24}
                                    & RPD    & -0.02 & \textbf{\textcolor{red}{1.00}} & -0.35 & \textbf{\textcolor{red}{1.00}} & \textbf{\textcolor{red}{1.00}} & \textbf{\textcolor{red}{0.84}} & \textbf{\textcolor{red}{1.00}} & \textbf{\textcolor{red}{0.19}} & \textbf{\textcolor{red}{0.61}} & \textbf{\textcolor{red}{0.07}} & -0.77 & \textbf{\textcolor{red}{0.19}} & \textbf{\textcolor{red}{0.53}} & \textbf{\textcolor{red}{0.37}} & \textbf{\textcolor{red}{1.00}} & \textbf{\textcolor{red}{0.55}} & \textbf{\textcolor{red}{0.14}} & \textbf{\textcolor{red}{1.00}} & \textbf{\textcolor{red}{0.08}} & -0.33 & -1.00 & \textbf{\textcolor{red}{0.05}} \\ \hline

        \end{tabular}

        \bigskip

        \setlength{\tabcolsep}{2pt} 
        \renewcommand{\arraystretch}{1.1} 
        \begin{tabular}{|l|l|l|l|l|l|l|l|l|l|l|l|l|l|l|l|l|l|l|l|l|l|l|l|}
        \hline
                                   &          & 22 & 23 & 24 & 25 & 26 & 27 & 28 & 29 & 30 & 31 & 32 & 33 & 34 & 35 & 36 & 37 & 38 & 39 & 40 & 41 & 42 & 43 \\ \hline
        \multirow{3}{*}{BC-RESNET-RNN} & SR - ORI & 0.87 & 0.00 & 0.03 & 0.15 & 0.33 & 1.00 & 0.65 & 0.45 & 0.58 & 0.85 & 0.75 & 0.18 & 0.92 & 0.73 & 1.00 & 0.00 & 0.02 & 0.15 & 0.08 & 0.07 & 0.12 & 0.45 \\ \cline{2-24}
                                    & SR - MOD & 0.72 & 0.02 & 0.00 & 0.07 & 0.52 & 0.95 & 0.40 & 0.07 & 0.65 & 0.23 & 0.35 & 0.00 & 0.63 & 0.00 & 1.00 & 0.00 & 0.00 & 0.00 & 0.00 & 0.02 & 0.00 & 0.03 \\ \cline{2-24}
                                    & RPD    & \textbf{\textcolor{red}{0.17}} & 0.00 & \textbf{\textcolor{red}{1.00}} & \textbf{\textcolor{red}{0.56}} & -0.55 & \textbf{\textcolor{red}{0.05}} & \textbf{\textcolor{red}{0.38}} & \textbf{\textcolor{red}{0.85}} & -0.11 & \textbf{\textcolor{red}{0.73}} & \textbf{\textcolor{red}{0.53}} & \textbf{\textcolor{red}{1.00}} & \textbf{\textcolor{red}{0.31}} & \textbf{\textcolor{red}{1.00}} & 0.00 & 0.00 & \textbf{\textcolor{red}{1.00}} & \textbf{\textcolor{red}{1.00}} & \textbf{\textcolor{red}{1.00}} & \textbf{\textcolor{red}{0.75}} & \textbf{\textcolor{red}{1.00}} & \textbf{\textcolor{red}{0.93}} \\ \hline
\multirow{3}{*}{BC-RESNET-T} & SR - ORI & 1.00 & 0.27 & 0.98 & 0.47 & 0.97 & 1.00 & 0.65 & 0.57 & 1.00 & 0.98 & 0.93 & 0.98 & 0.95 & 0.85 & 1.00 & 0.47 & 0.65 & 0.60 & 0.48 & 0.90 & 0.70 & 0.53 \\ \cline{2-24}
                                    & SR - MOD & 0.97 & 0.33 & 0.95 & 0.60 & 0.90 & 1.00 & 0.38 & 0.80 & 0.90 & 0.42 & 0.75 & 0.88 & 0.98 & 0.07 & 1.00 & 0.52 & 1.00 & 0.78 & 0.00 & 0.90 & 0.30 & 0.30 \\ \cline{2-24}
                                    & RPD    & \textbf{\textcolor{red}{0.03}} & -0.25 & \textbf{\textcolor{red}{0.03}} & -0.29 & \textbf{\textcolor{red}{0.07}} & 0.00 & \textbf{\textcolor{red}{0.41}} & -0.41 & \textbf{\textcolor{red}{0.10}} & \textbf{\textcolor{red}{0.58}} & \textbf{\textcolor{red}{0.20}} & \textbf{\textcolor{red}{0.10}} & -0.04 & \textbf{\textcolor{red}{0.92}} & 0.00 & -0.11 & -0.54 & -0.31 & \textbf{\textcolor{red}{1.00}} & 0.00 & \textbf{\textcolor{red}{0.57}} & \textbf{\textcolor{red}{0.44}} \\ \hline
\multirow{3}{*}{BC-VIT-T} & SR - ORI & 0.98 & 0.50 & 0.93 & 0.48 & 0.95 & 0.98 & 0.82 & 0.72 & 0.98 & 1.00 & 0.65 & 0.95 & 0.85 & 0.95 & 1.00 & 0.38 & 0.80 & 0.72 & 0.37 & 0.77 & 0.65 & 0.57 \\ \cline{2-24}
                                    & SR - MOD & 1.00 & 0.33 & 1.00 & 0.62 & 0.93 & 1.00 & 0.30 & 0.57 & 0.95 & 0.72 & 0.87 & 0.85 & 0.87 & 0.83 & 1.00 & 0.67 & 0.82 & 0.77 & 0.00 & 0.77 & 0.88 & 0.87 \\ \cline{2-24}
                                    & RPD    & -0.02 & \textbf{\textcolor{red}{0.33}} & -0.07 & -0.28 & \textbf{\textcolor{red}{0.02}} & -0.02 & \textbf{\textcolor{red}{0.63}} & \textbf{\textcolor{red}{0.21}} & \textbf{\textcolor{red}{0.03}} & \textbf{\textcolor{red}{0.28}} & -0.33 & \textbf{\textcolor{red}{0.11}} & -0.02 & \textbf{\textcolor{red}{0.12}} & 0.00 & -0.74 & -0.02 & -0.07 & \textbf{\textcolor{red}{1.00}} & 0.00 & -0.36 & -0.53 \\ \hline
\multirow{3}{*}{OpenVLA} & SR - ORI & 1.00 & 0.90 & 1.00 & 0.45 & 0.95 & 1.00 & 0.65 & 0.95 & 0.75 & 0.40 & 0.60 & 0.95 & 0.85 & 0.85 & 0.90 & 1.00 & 0.35 & 1.00 & 0.00 & 1.00 & 0.70 & 0.65 \\ \cline{2-24}
                                    & SR - MOD & 1.00 & 0.55 & 0.65 & 0.35 & 0.85 & 1.00 & 0.70 & 0.65 & 0.15 & 0.05 & 0.50 & 0.60 & 0.80 & 0.95 & 0.00 & 0.65 & 0.35 & 0.30 & 0.00 & 0.30 & 1.00 & 0.00 \\ \cline{2-24}
                                    & RPD    & 0.00 & \textbf{\textcolor{red}{0.39}} & \textbf{\textcolor{red}{0.35}} & \textbf{\textcolor{red}{0.22}} & \textbf{\textcolor{red}{0.11}} & 0.00 & -0.08 & \textbf{\textcolor{red}{0.32}} & \textbf{\textcolor{red}{0.80}} & \textbf{\textcolor{red}{0.87}} & \textbf{\textcolor{red}{0.17}} & \textbf{\textcolor{red}{0.37}} & \textbf{\textcolor{red}{0.06}} & -0.12 & \textbf{\textcolor{red}{1.00}} & \textbf{\textcolor{red}{0.35}} & 0.00 & \textbf{\textcolor{red}{0.70}} & 0.00 & \textbf{\textcolor{red}{0.70}} & -0.43 & \textbf{\textcolor{red}{1.00}} \\ \hline
\multirow{3}{*}{MaIL} & SR - ORI & 0.98 & 0.82 & 0.93 & 0.27 & 0.38 & 0.68 & 0.80 & 0.90 & 0.55 & 0.85 & 0.25 & 0.97 & 0.18 & 0.95 & 1.00 & 0.57 & 0.85 & 0.62 & 0.93 & 0.53 & 0.93 & 0.10 \\ \cline{2-24}
                                    & SR - MOD & 0.97 & 0.20 & 0.92 & 0.55 & 0.17 & 0.07 & 0.55 & 0.42 & 0.02 & 0.73 & 0.27 & 0.93 & 0.77 & 0.35 & 0.53 & 0.58 & 0.33 & 0.38 & 0.00 & 0.53 & 0.20 & 0.17 \\ \cline{2-24}
                                    & RPD    & \textbf{\textcolor{red}{0.02}} & \textbf{\textcolor{red}{0.76}} & \textbf{\textcolor{red}{0.02}} & -1.06 & \textbf{\textcolor{red}{0.57}} & \textbf{\textcolor{red}{0.90}} & \textbf{\textcolor{red}{0.31}} & \textbf{\textcolor{red}{0.54}} & \textbf{\textcolor{red}{0.97}} & \textbf{\textcolor{red}{0.14}} & -0.07 & \textbf{\textcolor{red}{0.03}} & -3.18 & \textbf{\textcolor{red}{0.63}} & \textbf{\textcolor{red}{0.47}} & -0.03 & \textbf{\textcolor{red}{0.61}} & \textbf{\textcolor{red}{0.38}} & \textbf{\textcolor{red}{1.00}} & 0.00 & \textbf{\textcolor{red}{0.79}} & -0.67 \\ \hline

        \end{tabular}
        \end{table*}

Table~\ref{tb:bm1_results_appendix} presents the success rates for 44 selected tasks, both unaffected by \pb (SR - ORI) and affected by \pb (SR - MOD), along with the Ratio Performance Delta (RPD). As shown in the table, all four baselines exhibit significant performance drops across most tasks, highlighting the negative impact of \pb.

\subsubsection{Numerical Results for BOSS-C2}
\begin{table*}[ht]
\centering
\caption{BOSS-C2 Results: Evaluation of Four Baselines on Two Sets of Modified Tasks (44 Tasks Each) with Two and Three Modifications Respectively.}
\label{tb:bm2_results_appendix}
\begin{tabular}{|c|c|c|c|c|c|}
\hline
                                 &                              & BC-RESNET-RNN & BC-RESNET-T & BC-VIT-T & OpenVLA \\ \hline
\multicolumn{1}{|l|}{Original}   & \multicolumn{1}{l|}{Avg. SR} & 0.43          & 0.83        & 0.84     & 0.81    \\ \hline
\multirow{2}{*}{1 Modification}  & Avg. SR                      & 0.25          & 0.67        & 0.74     & 0.54    \\ \cline{2-6} 
                                 & Avg. RPD                      & 0.67          & 0.35        & 0.34     & 0.54    \\ \hline
\multirow{2}{*}{2 Modifications} & Avg. SR                      & 0.25          & 0.55        & 0.53     & 0.46    \\ \cline{2-6} 
                                 & Avg. RPD                      & 0.64          & 0.50        & 0.55     & 0.68    \\ \hline
\multirow{2}{*}{3 Modifications} & Avg. SR                      & 0.23          & 0.49        & 0.56     & 0.35    \\ \cline{2-6} 
                                 & Avg. RPD                      & 0.65          & 0.59        & 0.50     & 0.74    \\ \hline
\end{tabular}
\end{table*}

Table~\ref{tb:bm2_results_appendix} presents the average success rates for 44 selected tasks, both unaffected by \pb (Original) and affected by \pb with varying numbers of modifications (1, 2, and 3). As shown in the table, all four baselines experience significant performance drops regardless of the number of modifications, underscoring the negative impact of \pb. Additionally, across all baselines, an increasing trend in RPD values is observed as the number of modifications increases.



\end{document}